\documentclass[acmsmall]{acmart}

\usepackage{colortbl}
\usepackage{xcolor}
\definecolor{Gray}{gray}{0.9}

\usepackage{tabularx}
\usepackage{multirow}
\usepackage{makecell}
\usepackage{caption}

\usepackage{url}
\usepackage{booktabs}
\usepackage{makecell}
\usepackage{xcolor} 
\usepackage{multirow}
\usepackage{enumerate}
\usepackage{pifont}
\usepackage{array}   
\usepackage{caption}
\usepackage{algorithm, algorithmic}
\usepackage{verbatim}
\usepackage{bbding}
\usepackage{bbm}
\usepackage{booktabs}
\usepackage{multicol}

\usepackage{colortbl} 

\usepackage{newfloat}
\usepackage{listings}

\usepackage{epsfig}
\usepackage{graphicx}
\usepackage{amsmath}

\usepackage{amssymb}

\usepackage{url}
\usepackage{booktabs}
\usepackage{makecell}
\usepackage{xcolor} 
\usepackage{multirow}
\usepackage{enumerate}
\usepackage{pifont}
\usepackage{array}   
\usepackage{caption}
\usepackage{algorithm, algorithmic}
\usepackage{verbatim}

\usepackage{makecell}
\usepackage{newfloat}
\usepackage{listings}

\usepackage{epsfig}
\usepackage{graphicx}
\usepackage{amsmath}

\usepackage{amssymb} 

\usepackage{url}
\usepackage{booktabs}
\usepackage{makecell}
\usepackage{xcolor} 
\usepackage{multirow}
\usepackage{enumerate}
\usepackage{pifont}
\usepackage{array}   
\usepackage{caption}
\usepackage{algorithm, algorithmic}
\usepackage{verbatim}
\usepackage{bbding}
\usepackage{bbm}
\usepackage{booktabs}
\usepackage{multicol}

\AtBeginDocument{%
  }

\setcopyright{acmlicensed}
\copyrightyear{2024}
\acmYear{2024}
\acmDOI{XXXXXXX.XXXXXXX}

\acmJournal{JACM}
\acmVolume{}
\acmNumber{}
\acmArticle{}
\acmMonth{5}

\begin{document}

\title{Mix-Modality Person Re-Identification: A New and Practical Paradigm}

\author{WEI LIU}
\email{liuwei@wust.edu.cn}
\affiliation{
\institution{School of Computer Science and Technology, Wuhan University of Science and Technology}
  \city{Wuhan}
  \country{China}
}

\author{XIN XU}
\email{xuxin@wust.edu.cn}
\affiliation{
    \institution{School of Computer Science and Technology, Wuhan University of Science and Technology}
  \city{Wuhan}
  \country{China}
}

\author{HUA CHANG}
\email{changhua@wust.edu.cn}
\affiliation{
    \institution{School of Computer Science and Technology, Wuhan University of Science and Technology}
  \city{Wuhan}
  \country{China}
}

\author{XIN YUAN}
\email{yuanxin@wust.edu.cn}
\affiliation{
    \institution{School of Computer Science and Technology, Wuhan University of Science and Technology}
  \city{Wuhan}
  \country{China}
}

\author{ZHENG WANG}
\email{wangzwhu@whu.edu.cn}
\affiliation{
  \institution{National Engineering Research Center for Multimedia Software, Institute of Artificial Intelligence, School of Computer Science, Wuhan University}
  \city{Wuhan}
  \country{China}
  }

\begin{abstract}
Current visible-infrared cross-modality person re-identification research has only focused on exploring the bi-modality mutual retrieval paradigm, and we propose a new and more practical mix-modality retrieval paradigm. Existing \textbf{V}isible-\textbf{I}nfrared person re-identification (VI-ReID) methods have achieved some results in the bi-modality mutual retrieval paradigm by learning the correspondence between visible and infrared modalities.
However, significant performance degradation occurs due to the modality confusion problem when these methods are applied to the new mix-modality paradigm.
Therefore, this paper proposes a \textbf{M}ix-\textbf{M}odality person re-identification (MM-ReID) task, explores the influence of modality mixing ratio on performance, and constructs mix-modality test sets for existing datasets according to the new mix-modality testing paradigm.
To solve the modality confusion problem in MM-ReID, we propose a \textbf{C}ross-\textbf{I}dentity \textbf{D}iscrimination \textbf{H}armonization \textbf{L}oss (CIDHL) adjusting the distribution of samples in the hyperspherical feature space, pulling the centers of samples with the same identity closer, and pushing away the centers of samples with different identities while aggregating samples with the same modality and the same identity. Furthermore, we propose a \textbf{M}odality \textbf{B}ridge \textbf{S}imilarity \textbf{O}ptimization \textbf{S}trategy (MBSOS) to optimize the cross-modality similarity between the query and queried samples with the help of the similar bridge sample in the gallery.
Extensive experiments demonstrate that compared to the original performance of existing cross-modality methods on MM-ReID, the addition of our CIDHL and MBSOS demonstrates a general improvement.
\end{abstract}



\received{20 February 2007}
\received[revised]{12 March 2009}
\received[accepted]{5 June 2009}

\maketitle

\begin{figure*}[!t]
\centering
\includegraphics[width=0.95\textwidth, height=0.5\textwidth]{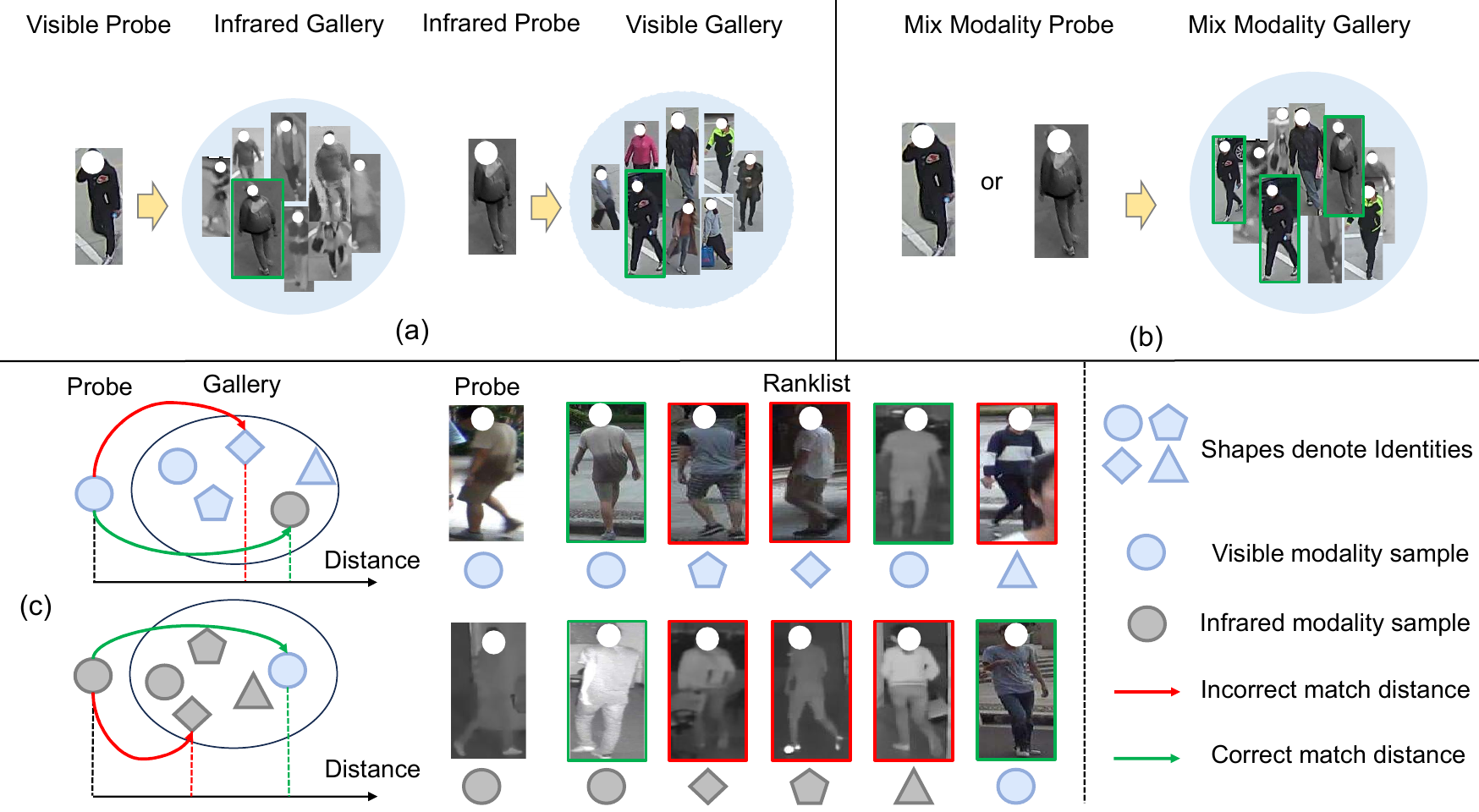}
\caption{(a) Existing bi-modality mutual retrieval test paradigms for VI-ReID use a visible probe image queried in an infrared image gallery or an infrared probe image queried in a visible image gallery. (b) Our proposed mix-modality testing paradigm for MM-ReID uses mix-modality probes to query in a mixed-modality gallery. (c) The unique challenge in MM-ReID is the interference of sample points with different identities in the same modality. In the figure, blue/gray represents visible/infrared modality samples, respectively, and different shapes represent different identities. It can be seen that due to more similar color and other identity-irrelevant information, the distance between samples of the same modality but different identities is closer than that between samples of the same identity with different modalities, which eventually leads to confusion during retrieval and reduces the accuracy.}
\label{fig:one}
\end{figure*}

\section{Introduction}


Person \textbf{Re}-\textbf{ID}entification (ReID) is a critical task in intelligent video security systems, aiming to retrieve specific pedestrians across a network of non-overlapping cameras~\cite{leng2019survey, ye2021deep, xu2022rank}. While \textbf{S}ingle-\textbf{M}odality ReID (SM-ReID) methods focusing on visible image retrieval have achieved significant advancements, they often fall short in low-light conditions where crime rates are notably higher. These illumination conditions lead to substantial information loss in RGB camera captures. To overcome this limitation, increasingly sophisticated cameras capable of automatically switching to infrared mode in low-light conditions are being integrated into video surveillance systems. This shift has spurred a growing research interest in \textbf{V}isible-\textbf{I}nfrared person re-identification (VI-ReID), which aims to tackle the challenge of cross-modality image matching~\cite{huang2022modality,liu2022learning,lu2020cross,ye2020dynamic}.



VI-ReID not only grapples with the common challenges of viewpoint, background, pose, and occlusion typical of SM-ReID task~\cite{cho2018pamm,wang2021horeid,xu2022towards,jin2020semantics, zhang2019densely, zhao2017spindle, ding2020multi} but also faces significant difficulties arising from modality differences. Despite considerable progress achieved in the bi-modality mutual retrieval paradigm, as illustrated in Fig~\ref{fig:one} (a) through learning potential correspondences between visible and infrared images, practical ReID scenarios pose additional complexities. As shown in Fig~\ref{fig:one} (b), pedestrian images in real applications may need to be identified across both day and night, necessitating a database that integrates both visible and infrared images—not merely a collection of one modality type. This integration often leads to a mix of what seems to be straightforward SM-ReID tasks into the existing cross-modality framework. However, as indicated in Fig~\ref{fig:one} (c), this approach results in a marked performance degradation, primarily due to the \textbf{`Modality Confusion'} problem. This issue stems from identity-independent features such as colors being more similar within the same modality, which confuses the matching of cross-modality identity information. More specifically, the impact of modality confusion on the current approach of learning only visible-infrared cross-modality correspondences is huge, due to the fact that there exists not only one correspondence from visible to infrared in the retrieval process, but also two same-modality correspondences, which is not taken into account by the existing approaches of VI-ReID. Under this influence, the natural similarity between samples of the same modality will disturb the perception of the algorithm, resulting in the distance between samples of different identities of the same modality being smaller than the distance between samples of the same identity of different modalities, and ultimately giving wrong retrieval results. To address these challenges, we introduce a new and practical \textbf{M}ix-\textbf{M}odality person re-identification (MM-ReID) task, creating mixed modality test sets for existing datasets and examining how the ratio of modality mixing affects retrieval performance.

To address the modality confusion challenge in MM-ReID, we introduce the \textbf{C}ross-\textbf{I}dentity \textbf{D}iscrimination \textbf{H}armonization \textbf{L}oss (CIDHL). This novel approach optimizes the sample distribution within a hyperspherical feature space. Specifically, CIDHL aggregates distances between samples of the same identity and modality towards their respective centers, while simultaneously drawing together centers of samples sharing identities across different modalities and distancing those of different identities, irrespective of modality. This strategy mitigates the effects of modality confusion. Additionally, we propose a \textbf{M}odality \textbf{B}ridge \textbf{S}imilarity \textbf{O}ptimization \textbf{S}trategy (MBSOS), which optimizes the cross-modality distance metric by identifying a similar bridge sample in the gallery to serve as an intermediary, thus refining the distance measures between the query and the queried samples. Extensive testing confirms that our methods—CIDHL and MBSOS—significantly enhance the model performance of existing cross-modality approaches, demonstrating their efficacy and adaptability in resolving modality confusion in MM-ReID.


The contributions of this paper are three aspects:
\begin{itemize}
\item \textbf{Paradigm Contribution:} We pioneer the \textbf{M}ix-\textbf{M}odality person re-identification (MM-ReID) task, introducing a novel testing paradigm that incorporates both visible and infrared modalities within a single framework.
\item \textbf{Empirical Contribution:} We investigate the impact of varying modality mixing ratios on cross-modality model performance and conclude the unique challenges of \textbf{`Modality Confusion'} specific to MM-ReID.
\item \textbf{Technical Contribution:} We introduce the \textbf{C}ross-\textbf{I}dentity \textbf{D}iscrimination Harmonization \textbf{L}oss (CIDHL) and the \textbf{M}odality \textbf{B}ridge \textbf{S}imilarity \textbf{O}ptimization \textbf{S}trategy (MBSOS). These transferable methodologies effectively address modality confusion, as validated by extensive empirical experiments.
\end{itemize}


\section{Related Work}
\subsection{Single-Modality Person Re-Identification}

\textbf{S}ingle-\textbf{M}odality person re-identification (SM-ReID) seeks to identify and retrieve all images of a specific pedestrian across a network of non-overlapping cameras using visible light images. With the advancement of deep learning technologies, SM-ReID has achieved substantial progress in recent years. However, the predominant focus of existing approaches has been on addressing changes in pose, viewpoint, and occlusion ~\cite{dou2022human, yuan2023searching}, with less attention given to the degradation of image quality in low-light conditions~\cite{min2017blind,min2018blind}, which are prevalent in high-crime areas. Some studies have explored low-light person re-identification, focusing on learning illumination-independent features and employing image enhancement techniques ~\cite{huang2019illumination,zeng2020illumination,zhang2020illumination,huang2020real, lu2023illumination}. Despite these efforts, the loss of pedestrian identity information due to the inherent limitations of RGB cameras in dimly lit environments remains a significant challenge~\cite{min2020study,min2021screen}. The deployment of advanced cameras capable of switching to infrared mode in low-light conditions has spurred further research~\cite{min2024exploring,min2024perceptual}. In this context, the \textbf{V}isible-\textbf{I}nfrared person re-identification (VI-ReID) task, which was first proposed by Song \textit{et al.}~\cite{wu2017rgb}, addresses the cross-modality matching of visible and infrared images to overcome these limitations.

\subsection{Visible-Infrared Person Re-Identification}
\textbf{V}isible-\textbf{I}nfrared person re-identification (VI-ReID) aims to utilize visible images to retrieve corresponding infrared images or vice versa. This bi-modality mutual search paradigm confronts not only the common challenges of single-modality ReID, such as changes in background, viewpoint, and pose but also the substantial challenge caused by discrepancies between visible and infrared images due to differing imaging technologies. Numerous deep learning strategies have been devised to mitigate both inter- and intra-modality variations. For instance, Wang \textit{et al.}~\cite{wang2019rgb} combined pixel and feature alignment to harmonize the feature distributions across modalities. Yet, they later recognized limitations in set-level alignment, leading to a refined approach that incorporates both set-level and instance-level alignment~\cite{wang2020cross}. Zhao \textit{et al.}~\cite{zhao2021joint} introduced random clothing color transformations to align features independently of color, addressing concerns that set-level alignment might overlook individual specifics.

Further, Fang \textit{et al.}~\cite{fang2023visible} pioneered semantic alignment by mapping semantically rich features to learnable prototypes, using affinity information to refine cross-modality identification. Traditional data augmentation techniques have also evolved within VI-ReID, with Ye \textit{et al.}~\cite{ye2021channel} developing a method to enhance image diversity by swapping and erasing color channels randomly. For localized feature extraction, Wei \textit{et al.}~\cite{wei2021flexible} utilized an adversarial learning framework with a flexible body partition model to segment and analyze different body parts. With the rise of Transformer technology, Liang \textit{et al.}~\cite{liang2023cross} adopted this approach for cross-modality ReID tasks, focusing on extracting highly discriminative features through modality-aware enhancement. In terms of modality transformation, Wei \textit{et al.}~\cite{wei2021syncretic} proposed synthesizing new modalities by merging features from both RGB and IR spectra. Meanwhile, Zhang \textit{et al.}~\cite{zhang2021towards} and Huang \textit{et al.}~\cite{huang2022modality} explored generating intermediate modalities to minimize differences, with Zhong \textit{et al.}~\cite{zhong2021grayscale} experimenting on converting infrared to visible images through colorization.

However, in a realistic application system of Re-ID, the queried pedestrians may appear both during daytime and nighttime, and the gallery to be queried also consists of data captured during the whole time period, \textit{i.e.}, the real application scenario requirement should be the retrieval of mix-modality images of specific pedestrians in a mix-modality gallery, rather than just a simple bi-modality mutual retrieval paradigm adopted by existing methods. Therefore, we propose a \textbf{M}ix-\textbf{M}odality person re-identification (MM-ReID) task according to the mix-modality paradigm.

\section{Methodology}
In this section, we elaborate on the novel \textbf{M}ix-\textbf{M}odality person re-identification (MM-ReID) paradigm and introduce two cornerstone methodologies: the \textbf{C}ross-\textbf{I}dentity \textbf{D}iscrimination \textbf{H}armonization \textbf{L}oss (CIDHL) and the \textbf{M}odality \textbf{B}ridge \textbf{S}imilarity \textbf{O}ptimization \textbf{S}trategy (MBSOS). More specifically, Section~\ref{subsec:MM-ReID} provides the task definition and formulation of our proposed MM-ReID together with an analysis of the unique modality confusion problem that exists in MM-ReID with respect to VI-ReID, Section~\ref{subsec:Motivation} describes the design motivation for the method and the differences with existing methods, Section~\ref{subsec:CIDHL} introduces the design idea and the detailed composition of the CIDHL loss, and Section~\ref{subsec:MBSOS} gives a detailed description of the steps and algorithmic procedure of the MBSOS strategy.

\subsection{Mix-Modality Person Re-identification}
\label{subsec:MM-ReID}
\subsubsection{\textbf{Task Definition and Formulation}} 
In this study, we use $\mathcal{Q}=\left\{\mathbf{q}_i \mid i=1,2, \ldots, N_{\mathrm{q}}\right\}$ to denote the query set, which consists of $N_{\mathrm{q}}$ probe images, and the gallery set as $\mathcal{G} = \left\{\mathbf{g}i \mid i = 1,2, \ldots, N{\mathrm{g}}\right\}$, containing $N_{\mathrm{g}}$ gallery images. As depicted in Figure~\ref{fig:one}, under the traditional bi-modality mutual retrieval paradigm of Visible-Infrared Re-Identification (VI-ReID), the sets $\mathcal{Q}$ and $\mathcal{G}$ each consist of images exclusively in one modality, either visible or infrared. Contrary to this, our proposed \textbf{M}ix-\textbf{M}odality Re-Identification (MM-ReID) paradigm includes images in both visible and infrared modalities within both $\mathcal{Q}$ and $\mathcal{G}$, addressing a more realistic and complex retrieval scenario.


\begin{figure}[!t]
\centering
\includegraphics[width=\textwidth, height=0.25\textwidth]{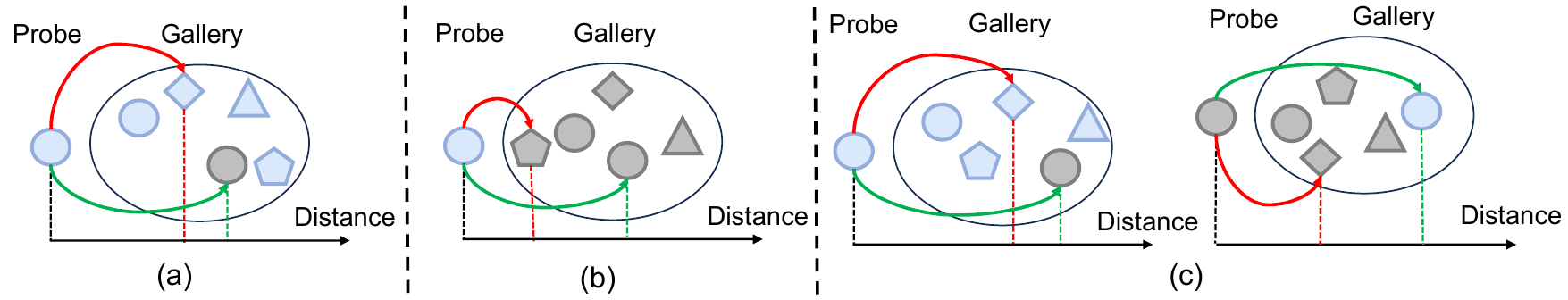}
\caption{An illustration of challenges in SM-ReID, VI-ReID, MM-ReID. Different geometries represent different identities, blue/gray represents visible/infrared modality samples, while green and red lines represent correct and incorrect matches. It can be seen that: (a) SM-ReID mainly faces the challenge of differences between different identities (shapes); (b) VI-ReID needs to face the challenge of modality (color) differences in addition to identity differences; and (c) MM-ReID needs to face the challenge of modality confusions (similar modalities possessing a closer proximity) in addition to identity and modality differences.}
\label{fig:two}
\end{figure}

\subsubsection{\textbf{Challenge Analyze}} 

As illustrated in Fig~\ref{fig:two}, the challenges inherent in \textbf{S}ingle-\textbf{M}odality ReID (SM-ReID) primarily stem from the need to accurately match pedestrian images across varied camera feeds, a task complicated by background noise, changes in viewpoint, and variations in pose. The complexity escalates in \textbf{V}isible-\textbf{I}nfrared ReID (VI-ReID), where the task involves matching images across the visible and infrared spectra, a process hampered by substantial differences in their imaging mechanisms. These modality differences represent significant obstacles to effective cross-modality matching. Furthermore, \textbf{M}ix-\textbf{M}odality ReID (MM-ReID) introduces additional complexities. Beyond the challenges faced in VI-ReID, MM-ReID contends with 'modality confusion,' where similarities irrelevant to identity, such as color consistency within a modality, can obscure vital identity-specific cues between the query ($\mathcal{Q}$) and gallery ($\mathcal{G}$) sets, thereby complicating the differentiation of individuals.


\begin{figure*}[!t]
\centering
\includegraphics[width=0.95\textwidth, height=0.45\textwidth]{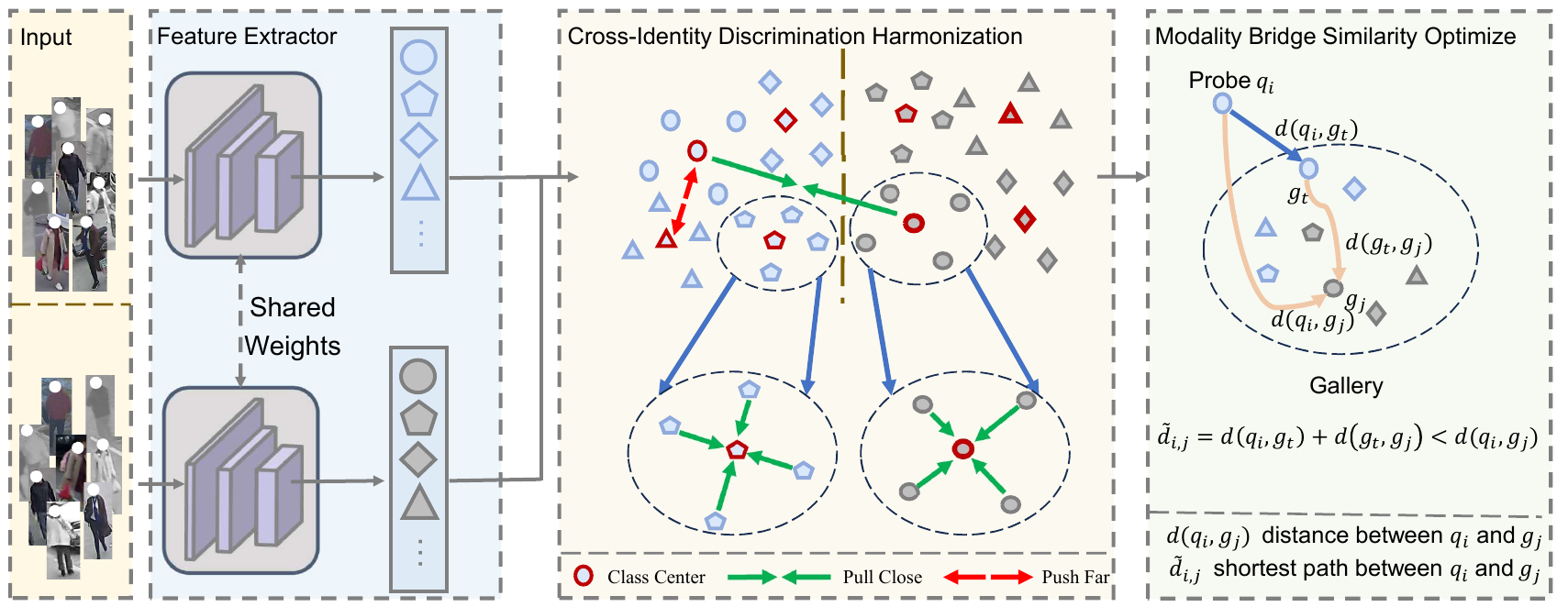}
\caption{An illustration of our proposed CIDHL and MBSOS. The mix-modality data is fed into two feature extractors with shared weights to extract the features, and under the constraints of CIDHL, the distance between the centers of the cross-modality same identity samples is pulled closer together pushing away the distance between the centers of the different identity samples of the same or different modalities, while at the same time pulling together the distance between the same identity sample point and the sample center. During the testing process, the extracted features are optimized by MBSOS to get the final shortest path with the help of bridge samples in the gallery set for obtaining the optimized distance metric $\Tilde{d}_{i,j}$.}
\label{fig:three}
\end{figure*}

\subsection{Method Design Motivation}
\label{subsec:Motivation}

To tackle the distinct modality confusion problem in the MM-ReID task, this paper introduces two innovative methods: \textbf{C}ross-\textbf{I}dentity \textbf{D}iscrimination \textbf{H}armonization \textbf{L}oss (CIDHL) and \textbf{M}odality \textbf{B}ridge \textbf{S}imilarity \textbf{O}ptimization \textbf{S}trategy (MBSOS). CIDHL specifically targets the confusion arising from similarity among same-modality samples by constraining these similarities in the metric space. This adjustment reduces the risk of confusing different-identity samples within the same modality when attempting to match cross-modality, same-identity positive samples.
Moreover, to mitigate modality confusion effects during the testing phase, MBSOS optimizes the similarity measure between cross-modality samples by leveraging same-modality samples in the gallery as a bridge. This is a departure from traditional VI-ReID methods, which typically focus only on constraining or exploiting the similarity across modalities. Our approach aims to refine both negative sample constraints and positive sample exploitation, specifically addressing the challenges posed by same-modality samples unique to the MM-ReID task.


\subsection{Cross-Identity Discrimination Harmonization Loss}
\label{subsec:CIDHL}

\subsubsection{\textbf{Feature Presentation Extractor}}
Our proposed methods, \textbf{C}ross-\textbf{I}dentity \textbf{D}iscrimination \textbf{H}armonization \textbf{L}oss (CIDHL) and \textbf{M}odality \textbf{B}ridge \textbf{S}imilarity \textbf{O}ptimization \textbf{S}trategy (MBSOS), are designed to seamlessly integrate with the feature extraction frameworks of existing VI-ReID methods. This integration ensures that there is no need for modifications to the baseline feature extraction processes of these methods.
As illustrated in Figure~\ref{fig:three}, we employ a generalized baseline feature representation extractor: a weight-shared, dual-branch convolutional neural network. This structure is adept at learning and extracting features from both visible and infrared images. The network’s effectiveness is further enhanced by the application of cross-entropy loss ($L_{\mathrm{id}}$) and our specifically developed CIDHL ($L_{\mathrm{CIDHL}}$), optimizing the feature extraction process and ensuring robust performance across varied modality inputs.
\subsubsection{\textbf{Triplet Loss Review}}
Introduced initially by FaceNet~\cite{schroff2015facenet}, triplet loss has become a cornerstone in identifying and enhancing relationships within small batches of samples in various machine learning applications. A significant advancement in this area is the development of the triplet hard loss~\cite{hermans2017defense}. This approach refines the basic triplet loss by dynamically selecting the most challenging samples within a mini-batch to maximize the informative value derived during training.
For a given mini-batch $X$, consisting of $P \times K$ images, where each of the $P$ identities is represented by $K$ images, triplet hard loss operates by selecting, for each anchor sample $x_{\mathrm{a}}$, the nearest negative sample $x_{\mathrm{n}}$ and the farthest positive sample $x_{\mathrm{p}}$. This selection strategy is designed to intensify the learning signals from the most informative triplets, involving the hardest positive and negative samples. The formulation of triplet hard loss, $L_{\mathrm{th}}$, effectively leverages these critical relationships within the batch to optimize the discriminative capability of the model, and the specific formula $L_{\mathrm{th}}$ can be expressed as follows:
\begin{equation}
\begin{aligned}
 L_{th}(X)={\sum_{i=1}^P \sum_{a=1}^K}[m+\overbrace{\max _{p=1 \ldots K}\left\|x_a^i-x_p^i\right\|_2}^{\text {hardest positive }} \\-
 \overbrace{\min _{\substack{j=1 \ldots P \\
n=1 \ldots K \\
j \neq i}}\left\|x_a^i-x_n^j\right\|_2}^{\text {hardest negative}}]_{+},
\end{aligned}
\label{eq1}
\end{equation}
where $x_{\mathrm{a}}^{i}$ denotes the $a-th$ image of $i-th$ identity pedestrian within a batch, $[x]_{+}$ stands for standard hinge loss, and $\left\|x_a^i-x_p^i\right\|_2$ denotes the Euclidean distance between $x_{\mathrm{a}}$ and $x_{\mathrm{p}}$, $m$ is the margin parameter for controling distance.

\subsubsection{\textbf{Cross-Identity Discrimination Harmonization} }
While the triplet hard loss provides a robust framework for learning from the most challenging samples, its effectiveness can be compromised by the presence of noisy data within the dataset. To counteract this issue, we have implemented a modification where the center of each identity is used as a substitute for individual samples. This approach helps stabilize the training process by reducing the influence of outliers or mislabeled instances.
For each identity, the center is calculated by averaging the features of all samples associated with that identity. This average is computed separately for each modality, ensuring that the specific characteristics of visible and infrared images are accurately represented. The identity centers are then utilized in place of individual samples when calculating triplet hard loss, thereby providing a more reliable basis for optimizing the model. This method not only mitigates the risk of steering the learning process in the wrong direction due to noisy data but also enhances the overall robustness of the model, and the centers for identities of different modalities are computed as follows:
\begin{equation}
\begin{aligned}
c_v^i & =\frac{1}{K} \sum_{j=1}^K v_j^i, \\
c_t^i & =\frac{1}{K} \sum_{j=1}^K t_j^i,
\end{aligned}
\end{equation}
where $v_{\mathrm{i}}^{j}$ and $t_{\mathrm{i}}^{j}$ denote the $j-th$ image of $i-th$ identity within visible and infrared modality, while the $c_{\mathrm{v}}^{i}$ and $c_{\mathrm{t}}^{i}$ represent the center of $i-th$ identity within two modalities.

Metric Learning in CIDHL: Enhancing Identity Discrimination Across Modalities
The fundamental principle of metric learning in CIDHL revolves around optimizing identity discrimination by manipulating distances in the feature space. The goal is to:
\begin{itemize}
    \item Minimize intra-identity distances: The model aims to decrease the distance between feature centers of the same identity, regardless of the modality. This step ensures that the same identity is represented more cohesively across different imaging conditions.
    \item Maximize inter-identity distances: Simultaneously, it is crucial to increase the distance between feature centers of different identities within the same modality and across different modalities. This separation helps distinguish individuals clearly.
    \item Aggregate intra-modality distances: To further refine the model's effectiveness, distances between samples of the same modality and identity are aggregated towards their respective centers. This aggregation aids in minimizing the modality-specific variations within the same identity, thus reducing the risk of boundary confusion where similar samples might be close to the decision boundaries.
\end{itemize}

The CIDHL adopts a structured approach, containing two parts, the first of which is targeted:
\begin{itemize}
    \item Cross-modality attraction: Distances between centers of the same identity across modalities are brought closer to foster a uniform identity representation across different sensory inputs.
    \item Intra- and inter-modality repulsion: Distances between centers of different identities, whether within the same modality or across modalities, are pushed apart to enhance discriminability.
\end{itemize}

The specific loss function of part one, denoted as $L_{\text{CID}}$, is formulated to dynamically adjust distances, ensuring an optimal balance between attraction and repulsion forces within the hyperspherical identity space, the $L_{cid}$ is calculated as:
\begin{equation}
\begin{aligned}
L_{cid}=&\sum_{i=1}^P\left[m+d(C_v^i-C_t^i)-\min _{n \in\{v, t\}, j \neq i}d(C_v^i-C_n^j)\right]_{+}\\
&+\sum_{i=1}^P\left[m+d(C_t^i-C_v^i)-\min _{n \in\{v, t\}, j \neq i}d(C_t^i-C_n^j)\right]_{+},
\label{eq3}
\end{aligned}
\end{equation}
where $C$ consisting of all the visible center $\left\{c_v^i \mid i=1, \cdots, P\right\}$ and $\left\{c_t^i \mid i=1, \cdots, P\right\}$ in the mini-batch $X$. Both two terms in the loss aim to learn cross-identity discrimination. 
Based on this, we further propose discrimination harmonization loss $L_{dh}$ for controlling the distance from the sample center of samples with the same modality and identity for solving the modality confusion problem due to the similarity of boundary samples. The $L_{dh}$ is calculated as follows:
\begin{equation}
\begin{aligned}
L_{dh}=&\left[m+\max _{1 \leq j \leq K}d(C_v^i-v_j^i)-\min _{1 \leq j \leq K}d(C_v^i-t_j^i)\right]_{+}+ \\
&\left[m+\max _{1 \leq j \leq K}d(C_t^i-t_j^i)-\min _{1 \leq j \leq K}d(C_t^i-v_j^i)\right].
\label{eq4}
\end{aligned}
\end{equation}
The final CIDHL $L_{CIDHL}$ can be calculated as:
\begin{equation}
L_{CIDHL}=L_{cid}+\delta L_{dh},
\label{eq5}
\end{equation}
where $\delta$ is the tradeoff parameter between $L_{cid}$ and $L_{dh}$.

\subsection{Modality Bridge Similarity Optimization Strategy}
\label{subsec:MBSOS}
The \textbf{M}odality \textbf{B}ridge \textbf{S}imilarity \textbf{O}ptimization \textbf{S}trategy (MBSOS) addresses the modality confusion issue by innovatively transforming the distance calculation process between a query sample and a gallery sample into a pathfinding problem. Here’s how it works:
\begin{itemize}
    \item Bridge Node Concept: Each query sample, $q_i \in \mathcal{Q}$, seeks the optimal path to a gallery sample, $g_j \in \mathcal{G}$, not by direct distance calculation but through intermediate "bridge nodes" within the gallery.
    \item Pathfinding Optimization: The strategy uses these bridge nodes, which are samples in the gallery of similar modality to the query, to create a modality-consistent path. This method effectively reduces the impact of direct modality differences on the distance metric.
    \item Distance Optimization: By navigating through these intermediate nodes, the system optimizes the apparent distance between $q_i$ and $g_j$. This optimization aims to provide a more accurate similarity measure by leveraging intra-modality consistency to guide cross-modality comparisons.
\end{itemize}
The core objective of MBSOS is to enhance the robustness of cross-modality person re-identification by minimizing the distortive effects of modality differences on the identity-matching process. This approach not only streamlines the matching process but also significantly improves the precision of identity verification across different imaging conditions.

\subsubsection{\textbf{Distance Map Construction}}
To obtain the relationship between sample points, we first calculate the distance $d(q_i, g_i)$ between query sample $q_{i}$ and gallery sample $g_{i}$, as well as the distance $d(g_i, g_j)$ between gallery samples $g_{i}$ and $g_{j}$ as follows:
\begin{equation}
    \begin{aligned}
        d(q_i,g_j)=&\left\|q_i-g_j\right\|_2\\
        d(g_j,g_t)=&\left\|q_j-g_t\right\|_2 ,
    \end{aligned}
\end{equation}
where$(i=1,2, \ldots, N_q)$ , $(j=1,2, \ldots, N_g)$ and $(t=1,2, \ldots, N_g)$. Then, we construct a distance map $\mathcal{M}_{QG}(\mathcal{V}, \mathcal{E})$ by $d(q_i, g_i)$ to represent the distances between the query set and gallery set, and another distance map $\mathcal{M}_{GG}(\mathcal{V}, \mathcal{E})$ using $d(g_j, g_t)$ to represent the distances between gallery samples in the gallery set. In $\mathcal{M}_{QG}$ and $\mathcal{M}_{GG}$, each vertex $\mathcal{V}$ represents a pedestrian image, and each edge $\mathcal{E}$  represents the distance between two vertices.
Considering that modality confusion is mainly caused by the fact that the same modality naturally has a closer distance, we propose an appropriate scaling adjustment for the distance of the edge $\mathcal{E}$ of the same modality to attenuate the natural effects of this imaging mechanism, the edges between query sample and gallery sample of the $\mathcal{M}_{QG}$ adjusted as follows:
\begin{equation}
{\mathcal{E}}_{\mathrm{qg}}^{i j}= \begin{cases}\lambda\mathrm{d(q_i, g_j)} & \text { if } \mathrm{ml(q_i)=ml(g_j)} \\ \mathrm{d(q_i, g_j)} & \text { otherwise, }\end{cases}
\label{eq7}
\end{equation}
where $\mathcal{E}_{\mathrm{qg}}^{i j}$ is the adjusted distance between $i-th$ and $j-th$ vertex in $\mathcal{M}_{QG}$. The $\mathrm{ml(q_i)}$ and $\mathrm{ml(g_j)}$ are the modality label of the $q_i$ and $g_j$. The edges in $\mathcal{M}_{GG}$ can be adjusted similarly as follows:
\begin{equation}
{\mathcal{E}}_{\mathrm{gg}}^{j t}= \begin{cases}\lambda\mathrm{d(g_j, g_t)} & \text { if } \mathrm{ml(g_i)=ml(g_t)} \\ \mathrm{d(g_j, g_t)} & \text { otherwise. }\end{cases}
\label{eq8}
\end{equation}

\begin{algorithm}
\renewcommand{\algorithmicrequire}{\textbf{Input:}}
    \renewcommand{\algorithmicensure}{\textbf{Output:}}
\caption{modality Bridge Similarity Optimization Strategy}
\begin{algorithmic}[1]
\label{MBSOS}
\REQUIRE Query set $\mathcal{Q}$, Gallery set $\mathcal{G}$, Query-gallery distance map $\mathcal{M}_{QG}$, gallery-gallery distance $\mathcal{M}_{GG}$, modality label $ml$
\ENSURE Optimized $\Tilde{\mathcal{M}}_{QG}$
\FOR{each $q_i$, $g_j$ in $\mathcal{M}_{QG}$}
    \IF{$ml(q_i)$=$ml(g_j)$}
        \STATE adjust $\mathcal{E}_{qg}^{ij}$ according to Eq.~\ref{eq7}.
    \ENDIF
\ENDFOR
\FOR{each $g_j$, $g_t$ in $\mathcal{M}_{GG}$}
    \IF{$ml(g_j)$=$ml(g_t)$}
        \STATE adjust $\mathcal{E}_{gg}^{jt}$ according to Eq.~\ref{eq8}.
    \ENDIF
\ENDFOR
\FOR{each $q_i$ in $\mathcal{Q}$ and $g_j$ in $\mathcal{G}$}
    \FOR{$g_t$ in $\mathcal{G}$}
            \STATE Calculate $p_{ij}^{t} = d(q_i, g_t, g_j)$ according to Eq.~\ref{eq9}.
    \ENDFOR
    \STATE Sort $p_{ij}^{t}$ for all $g_t$.
    \STATE Calculate $\Tilde{d}_{ij}$ according to Eq.~\ref{eq10}.
\ENDFOR
\STATE Construct $\Tilde{\mathcal{M}}_{QG}$ by all the $\Tilde{d}_{ij}$
\RETURN $\Tilde{\mathcal{M}}_{QG}$
\end{algorithmic}
\end{algorithm}

\subsubsection{\textbf{Modality Brige Optimization}}
After constructing a distance graph $\mathcal{M_{QG}}$, 
computing the nearest distance $\Tilde{d_{ij}}$ from a given probe $\mathrm{q_i}$ to a point $\mathrm{g_j}$ in the gallery can be viewed as finding the shortest path from a probe node $\mathrm{q_i}$ to a gallery node $\mathrm{g_j}$ in the $\mathcal{M_{QG}}$ with the help of a bridge node $g_t$ in $\mathcal{M}_{GG}$. More specifically, a path $\mathrm{p_{ij}^{t}}$ from a probe node $\mathrm{q_i}$ to a gallery node $\mathrm{g_j}$ in the $\mathcal{M_{QG}}$ can be represented as follows:
\begin{equation}
    p_{ij}^{t} = (q_i,g_t,g_j)
           = \mathcal{E}_{qg}^{it}+\mathcal{E}_{gg}^{tj}.
\label{eq9}
\end{equation}
Then, the set of all paths between $\mathrm{q_i}$ and $\mathrm{g_j}$ is $\mathcal{P} = (p_{ij}^{1},\dots, p_{ij}^{n})$, where $n = N_G$. Therefore, the shortest distance $\Tilde{d}_{ij}$ between $q_i$ and $g_j$ can be calculated as follows:
\begin{equation}
\Tilde{d}_{ij}=\min _{1 \leq t \leq N_g} \mathcal{P} =\min _{1 \leq t \leq N_g}\left\{\mathcal{E}_{qg}^{it}+\mathcal{E}_{gg}^{tj}\right\}.
\label{eq10}
\end{equation}
Finally, we obtain the optimized distance map $\Tilde{\mathcal{M}}_{QG}$ constructed by all the shortest distance $\Tilde{d}_{ij}$ between the query nodes $q_i$ and $g_j$, which is optimized with the help of a bridge node $g_t$. It is worth mentioning that in some cases, the direct distance from $q_i$ to $g_j$ is the shortest distance without the help of any intermediate bridge node. The algorithmic procedure of MBSOS is described in Algorithm~\ref{MBSOS}.

\begin{table}[]
    \centering
        \caption{The comparison of three datasets.}
    \begin{tabular}{cccc}
    \hline Datasets & SYSU-MM01 & RegDB & LLCM \\
\hline ID\_number & 491 & 412 & 1,064 \\
\hline Train\_ID\_number & 395 & 206 & 713 \\
\hline Test\_ID\_number & 96 & 206 & 351 \\
\hline Visible image & 287,628 & 4,120 & 25,626 \\
\hline Infrared image & 15,792 & 4,120 & 21,141 \\
\hline Total image & 303,420 & 8,240 & 46,767 \\
\hline
    \end{tabular}
    \label{dataset}
\end{table}

\section{Experimental Design and Results}
In this section we first present the adopted dataset, evaluation metrics, mix-modality paradigm, and method implementation details in Section~\ref{Datasets}. In Section~\ref{Comparison} the original performance of the existing methods and the performance with the addition of our CIDHL and MBSOS are compared. In Section~\ref{Ablation} we do ablation experiments on the hyperparameters $\delta$ and $\lambda$. Finally, in Section~\ref{Visualization} we show the visualization results of the AGW algorithm with the addition of our CIDHL after the t-SNE dimensionality reduction.

\begin{figure*}[!t]
\centering
\includegraphics[width=0.95\textwidth, height=0.5\textwidth]{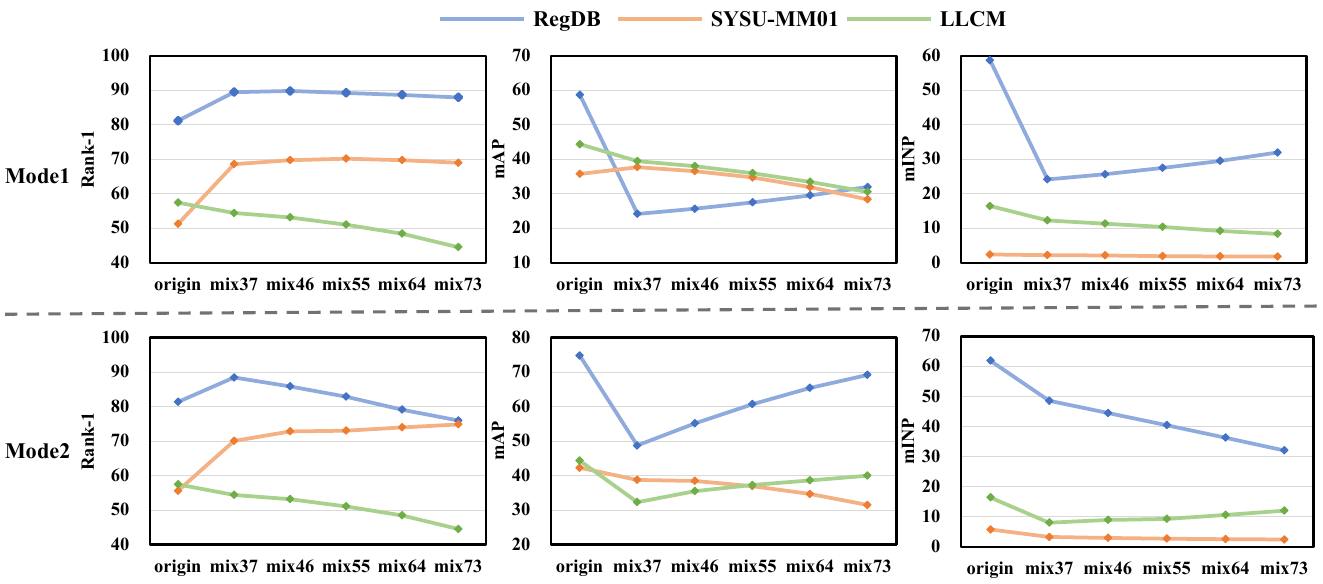}
\caption{The effect of different modality mixing ratios with respect to the AGW model performance in Rank-1, mAP, and mINP metrics on the three datasets. It can be seen that a general performance degradation arises on the other metrics except for the Rank-1 metrics that rise on some of the datasets. The origin stands for the unmixed dataset, mix37 stands for the query set to the gallery set with a ratio of $3:7$ for the visible images and infrared images, and so forth.}
\label{fig:four}
\end{figure*}

\subsection{Datasets and Settings}
\label{Datasets}
\subsubsection{\textbf{Datasets}}
As shown in Table~\ref{dataset}, we employed three public datasets RegDB~\cite{nguyen2017person}, SYSU-MM01~\cite{wu2017rgb}, and LLCM~\cite{zhang2023diverse}. The RegDB, SYSU-MM01, and LLCM datasets are prominent in the field of cross-modal person re-identification, utilizing dual-camera setups for data collection in visible and infrared modalities. RegDB includes $412$ pedestrians with $10$ visible and $10$ thermal infrared images each, totaling $4,120$ images per modality. It offers two testing modes: infrared image search using visible images and vice versa. SYSU-MM01, a larger dataset, contains $287,628$ RGB and $15,792$ IR images across $491$ IDs, featuring operational modes like All-Search and Indoor-Search in both single-shot and multi-shot formats. The LLCM dataset, the largest to date for VI-ReID, captured using nine cameras, comprises $46,767$ images across $1,064$ identities, split roughly into training and testing sets with two search modes similar to RegDB. Each dataset facilitates rigorous evaluation in cross-modal analysis, maintaining its integrity in segmenting data for varied testing scenarios.

\subsubsection{\textbf{Metrics}}
 The cumulative matching characteristics (CMC)~\cite{moon2001computational} at Rank-1, Rank-5, Rank-10, mean average precision (mAP) and mean inverse negative penalty (mINP)~\cite{ye2021deep} 
are adopted as evaluation metrics in this paper.

\subsubsection{\textbf{Mix-Modality Paradigm}}
We propose a new mix-modality paradigm by assigning the visible/infrared images of the same pedestrian in the test set to the query set and the gallery set, respectively, according to the modality mixing ratio. For example, if the modality mixing ratio is $3:7$, $3/10ths$ of the visible images and $7/10ths$ of the infrared images of the same pedestrian are constructed into the query set, and the remaining images are composed into the gallery set.

As shown in Figure~\ref{fig:three}, under such a testing paradigm, we tested the performance of the AGW model on the RegDB, SYSU-MM01, and LLCM datasets in the Rank-1, mAP, and mINP metrics. It can be seen that the performance of the model decreases in all other metrics and mixing ratios, except for the Rank-1 metric, which shows an improvement in the RegDB and SYSU-MM01 datasets. We analyze that the reason for the performance improvement in these two datasets is caused by the presence of image pairs with the same identity and same modality that are extremely easy to match and that the decrease in the other metrics is due to modality confusion that leads to a decrease in the overall recognition ability of the model. It is worth mentioning that, in order to measure the overall performance of the model, we adopt the multi-shot all way of testing, \textit{i.e.}, all the images of the target pedestrians will be included in the gallery set, and it can be seen that in this case, in the large-scale dataset, the model's mINP metrics appear to be greatly degraded because the query samples are farther away from the hardest positive samples than the majority of the negative samples.

\subsubsection{\textbf{Implementation Details}}
All experiments were conducted with the PyTorch framework and an RTX3090 GPU. Limited by the length and open source status of the algorithms, we conducted experiments with DART~\cite{yang2022learning}, LUPI~\cite{alehdaghi2022visible}, AGW~\cite{ye2021deep}, HCTL~\cite{liu2020parameter}, and DEEN~\cite{zhang2023diverse}, continuing the basic settings of these methods for training. The $m$ in Eq.1, Eq.3, and Eq.4 is set to $0.3$, the $\delta$ in Eq.5 is set to $0.2$, the $\lambda$ in Eq.7 and Eq.8 is set to $0.999$.

\begin{table}[]
\caption{Comparison to the state-of-the-art methods on RegDB datasets in mode 1. The \textbf{bold} result is the best of the same method, while the \textcolor{red}{$\uparrow$}is the method with performance improvement (the same as below).}
\label{tab:1}
\begin{tabular}{c|ccccc}
\hline
\multirow{2}{*}{Method}       & \multicolumn{5}{c}{Mode 1}    \\ \cline{2-6} 
                              & Rank-1 & Rank-5 & Rank-10 & mAP &mINP \\ \hline
DART $w/o$ mix       &72.80   &82.04    &86.74    &68.35 &56.07   \\
DART                 &98.52   &99.74    &99.74    &81.51 &87.63  \\ \hline
AGW $w/o$ mix       &81.23   &90.39    &93.93&74.81 &58.75   \\
AGW        &89.34   &96.21    &98.19&60.78 &27.57  \\
AGW $+$ M                     &81.83   &90.56    &94.44&64.45\textcolor{red}{$\uparrow$} &\textbf{39.46}\textcolor{red}{$\uparrow$}  \\
AGW $+$ C                     &\textbf{95.43}\textcolor{red}{$\uparrow$}   &\textbf{98.40}\textcolor{red}{$\uparrow$}    &\textbf{99.20}\textcolor{red}{$\uparrow$}&\textbf{69.62}\textcolor{red}{$\uparrow$}   &35.78\textcolor{red}{$\uparrow$}     \\
AGW $+$ C $+$ M               &94.34\textcolor{red}{$\uparrow$}   &98.02\textcolor{red}{$\uparrow$}    &99.03\textcolor{red}{$\uparrow$}& 68.42\textcolor{red}{$\uparrow$}  &34.04 \textcolor{red}{$\uparrow$}    \\ \hline
HCTL $w/o$ mix &88.90   &94.47    &96.18&80.67   &66.13     \\
HCTL  &78.63   &86.69    &90.30&59.35   &21.38     \\
HCTL $+$ M                    &77.93   &87.48\textcolor{red}{$\uparrow$}    &91.60\textcolor{red}{$\uparrow$}&58.66    &\textbf{22.72}\textcolor{red}{$\uparrow$}     \\
HCTL $+$ C                    &\textbf{90.05}\textcolor{red}{$\uparrow$}   &\textbf{96.21}\textcolor{red}{$\uparrow$}    &\textbf{98.32}\textcolor{red}{$\uparrow$}&\textbf{61.81}\textcolor{red}{$\uparrow$}   &21.38      \\
HCTL $+$ C $+$ M              &78.29   &86.76\textcolor{red}{$\uparrow$}    &91.42\textcolor{red}{$\uparrow$}&56.81    &19.42     \\\hline
DEEN $w/o$ mix &87.95   &93.89  &96.24&80.74     &65.64    \\
DEEN  &97.55   &99.33    &99.69&78.24  &51.58       \\
DEEN $+$ M                    &97.58\textcolor{red}{$\uparrow$}   &99.31    &99.71\textcolor{red}{$\uparrow$}&78.17    &51.66\textcolor{red}{$\uparrow$}      \\
DEEN $+$ C                    &\textbf{97.82}\textcolor{red}{$\uparrow$}   &99.35\textcolor{red}{$\uparrow$}    &99.71\textcolor{red}{$\uparrow$} &79.59\textcolor{red}{$\uparrow$}    &53.22\textcolor{red}{$\uparrow$}     \\
DEEN $+$ C $+$ M              &97.90\textcolor{red}{$\uparrow$}   &\textbf{99.39}\textcolor{red}{$\uparrow$}    &\textbf{99.72}\textcolor{red}{$\uparrow$}  &\textbf{79.82}\textcolor{red}{$\uparrow$}     &\textbf{53.79}\textcolor{red}{$\uparrow$}    \\ \hline

\end{tabular}
\end{table}

\begin{table}[]
\caption{Comparison to the state-of-the-art methods on RegDB datasets in mode 2.}
\label{tab:2}
\begin{tabular}{c|ccccc}
\hline
\multirow{2}{*}{Method}       & \multicolumn{5}{c}{Mode 2}    \\ \cline{2-6} 
                              & Rank-1 & Rank-5 & Rank-10 & mAP &mINP \\ \hline
DART $w/o$ mix       &88.95   &94.17    &96.32    &84.19 &74.67   \\
DART                 &98.26   &99.59    &99.87    &81.34 &56.67   \\ \hline 
AGW $w/o$ mix        &81.44   &90.16    &93.82&75.96 &61.91  \\
AGW         &82.93   &91.72    &95.41&65.42  &40.49 \\
AGW $+$ M                     &89.42\textcolor{red}{$\uparrow$}    &96.36\textcolor{red}{$\uparrow$}     &98.52\textcolor{red}{$\uparrow$} &61.08  &27.87 \\
AGW $+$ C                     &94.19\textcolor{red}{$\uparrow$}    &98.14\textcolor{red}{$\uparrow$}     &99.20\textcolor{red}{$\uparrow$} &68.53\textcolor{red}{$\uparrow$}    &33.93\\
AGW $+$ C $+$ M               &\textbf{95.49}\textcolor{red}{$\uparrow$}    &\textbf{98.58}\textcolor{red}{$\uparrow$}     &\textbf{99.24}\textcolor{red}{$\uparrow$} &\textbf{69.59}\textcolor{red}{$\uparrow$}   &\textbf{36.05}      \\ \hline
HCTL $w/o$ mix  &90.95   &95.28    &97.08&82.43    &68.84    \\
HCTL  &78.08   &86.66    &90.98&58.79   &22.50     \\
HCTL $+$ M                    &79.60\textcolor{red}{$\uparrow$}    &87.80\textcolor{red}{$\uparrow$}     &91.24\textcolor{red}{$\uparrow$} &59.51\textcolor{red}{$\uparrow$}  &21.35        \\
HCTL $+$ C                    &78.91\textcolor{red}{$\uparrow$}    &87.40\textcolor{red}{$\uparrow$}     &91.55\textcolor{red}{$\uparrow$} &56.61   &\textbf{25.27}\textcolor{red}{$\uparrow$}      \\
HCTL $+$ C $+$ M              &\textbf{90.10}\textcolor{red}{$\uparrow$}    &\textbf{96.60}\textcolor{red}{$\uparrow$}     &\textbf{98.70}\textcolor{red}{$\uparrow$} &\textbf{61.87}\textcolor{red}{$\uparrow$}     &20.00     \\\hline
DEEN $w/o$ mix &90.15   &95.42    &97.28&82.84  &69.18       \\
DEEN  &97.44   &99.19    &99.61&80.28     &54.88    \\
DEEN $+$ M                    &97.47\textcolor{red}{$\uparrow$}    &99.17    &99.67\textcolor{red}{$\uparrow$} &80.74\textcolor{red}{$\uparrow$}     &55.92\textcolor{red}{$\uparrow$}      \\
DEEN $+$ C                    &97.36   &99.19\textcolor{red}{$\uparrow$}     &99.69\textcolor{red}{$\uparrow$}  &80.82\textcolor{red}{$\uparrow$}    &55.62\textcolor{red}{$\uparrow$}      \\
DEEN $+$ C $+$ M              &\textbf{97.63}\textcolor{red}{$\uparrow$}    &\textbf{99.34}\textcolor{red}{$\uparrow$}     &\textbf{99.70}\textcolor{red}{$\uparrow$}   &\textbf{81.53}\textcolor{red}{$\uparrow$}   &\textbf{56.88}\textcolor{red}{$\uparrow$}       \\ \hline
\end{tabular}
\end{table}

\begin{table}[]
\caption{Comparison to the state-of-the-art methods on SYSU-MM01 datasets in mode 1.}
\label{tab:3}
\begin{tabular}{c|ccccc}
\hline
\multirow{2}{*}{Method}       & \multicolumn{5}{c}{Mode 1}    \\ \cline{2-6} 
                              & Rank-1 & Rank-5 & Rank-10 & mAP &mINP \\ \hline
DART $w/o$ mix       &55.31   &82.66    &90.54    &52.80 &38.42   \\
DART                 &72.62   &91.15    &95.41    &43.18 &7.84   \\
LUPI $w/o$ mix       &45.64   &76.85    &86.28    &46.75 &34.95   \\
LUPI                 &53.99   &80.87    &88.27    &31.82 &4.63   \\ \hline
AGW $w/o$ mix                   &51.33   &80.31    &88.59&35.82 &2.43  \\ 
AGW         &70.22   &91.76   &95.99&34.78 &1.97   \\
AGW $+$ M                     &70.73\textcolor{red}{$\uparrow$}   &92.02\textcolor{red}{$\uparrow$}    & 96.09\textcolor{red}{$\uparrow$}&34.76 &1.96   \\
AGW $+$ C                     &88.76\textcolor{red}{$\uparrow$}   &97.36\textcolor{red}{$\uparrow$}   &98.84\textcolor{red}{$\uparrow$} &\textbf{48.79}\textcolor{red}{$\uparrow$} &3.29\textcolor{red}{$\uparrow$}   \\
AGW $+$ C $+$ M               &\textbf{89.08}\textcolor{red}{$\uparrow$} &\textbf{97.53}\textcolor{red}{$\uparrow$}    &\textbf{98.97}\textcolor{red}{$\uparrow$} &48.54\textcolor{red}{$\uparrow$}  &\textbf{3.30}\textcolor{red}{$\uparrow$}  \\ \hline
HCTL $w/o$ mix  &69.55 &90.01    &95.95     &49.98  &7.26 \\
HCTL  &60.93  &84.63    &90.47     &30.30  &1.80  \\
HCTL $+$ M                    &61.18\textcolor{red}{$\uparrow$}  &84.77\textcolor{red}{$\uparrow$}  &90.50\textcolor{red}{$\uparrow$}&30.61\textcolor{red}{$\uparrow$}   &1.87\textcolor{red}{$\uparrow$}      \\
HCTL $+$ C                    &\textbf{84.92}\textcolor{red}{$\uparrow$}  &96.66\textcolor{red}{$\uparrow$}   &98.65\textcolor{red}{$\uparrow$}&\textbf{43.21}\textcolor{red}{$\uparrow$}  &2.56\textcolor{red}{$\uparrow$}       \\
HCTL $+$ C $+$ M              &85.51\textcolor{red}{$\uparrow$}  &\textbf{96.73}\textcolor{red}{$\uparrow$} &\textbf{98.69}\textcolor{red}{$\uparrow$}  &43.11\textcolor{red}{$\uparrow$}      &\textbf{2.58}\textcolor{red}{$\uparrow$}   \\\hline
DEEN $w/o$ mix &66.53  &89.72  &96.27  &11.04  &6.43       \\
DEEN  &76.87  &93.60  &96.89  &37.01    &2.63     \\
DEEN $+$ M                    &60.42  &83.84  &90.92  &29.29     &2.73\textcolor{red}{$\uparrow$}     \\
DEEN $+$ C                    &\textbf{80.63}\textcolor{red}{$\uparrow$}  &\textbf{94.43}\textcolor{red}{$\uparrow$}  &97.06\textcolor{red}{$\uparrow$}  &\textbf{47.04}\textcolor{red}{$\uparrow$}     &3.51\textcolor{red}{$\uparrow$}    \\
DEEN $+$ C $+$ M              &79.87\textcolor{red}{$\uparrow$}  &94.19\textcolor{red}{$\uparrow$}  &\textbf{97.13}\textcolor{red}{$\uparrow$}  &\textbf{47.04}\textcolor{red}{$\uparrow$}    &\textbf{3.76}\textcolor{red}{$\uparrow$}  \\ \hline
\end{tabular}
\end{table}

\subsection{Comparison with State-of-the-Arts}
\label{Comparison}
We evaluate the performance of our proposed method on AGW, HCTL, DEEN, and three datasets, RegDB, SYSU-MM01, and LLCM in Rank-1, Rank-5, Rank-10, mAP, and mINP. Where Mode 1 represents that we input the query set to the IR branch of the original model and the gallery set to the visible branch, and Mode 2 represents the opposite. The ``$w/o$ mix'' denotes the performance on the origin bi-modality mutual retrieval paradigm while others are on our mix-modality paradigm. The ``$+$M'', ``$+$C'', and ``$+$C$+$M'' denote the performance of the method added MBSOS, CIDHL, and both, respectively.

\textbf{Comparison on RegDB.}
As can be seen in Tables~\ref{tab:1} and~\ref{tab:2}, a general improvement in performance occurs with the addition of our method. In Mode 1, the AGW algorithm achieves the best performance with the addition of CIDHL, showing ``$\mathbf{+6.09\%}$'', ``$\mathbf{+2.19\%}$'', ``$\mathbf{+1.01\%}$'', ``$\mathbf{+8.84\%}$'', and ``$\mathbf{+8.21\%}$'' improvement in Rank-1, Rank-5, Rank-10, mAP, and mINP metrics, respectively.  The addition of MBSOS also shows a performance improvement of ``$\mathbf{+3.67\%}$'' and ``$\mathbf{+11.89\%}$'' in the mAP and mINP metrics, which reflects the overall performance of the model, despite a slight decrease in the rank metric. In Mode 2, the AGW algorithm with the addition of both CIDHL and MBSOS achieves the best performance, showing ``$\mathbf{+11.42\%}$'', ``$\mathbf{+9.52\%}$'', ``$\mathbf{+8.02\%}$'', ``$\mathbf{+2.46\%}$'', and ``$\mathbf{+2.00\%}$'' improvements in Rank-1, Rank-5, Rank-10, mAP, and mINIP metrics, respectively. Unlike Model 1 the performance of the model with the addition of MBSOS improves on the Rank metric. The performance of other algorithms is similar to AGW.

\textbf{Comparison on SYSU-MM01.}
As can be seen in Tables~\ref{tab:3} and~\ref{tab:4}, a general improvement in performance occurs with the addition of our method on SYSU-MM01, too. In Mode 1, the AGW algorithm achieves the best performance with the addition of both CIDHL and MBSOS, showing ``$\mathbf{+18.86\%}$'', ``$\mathbf{+5.76\%}$'', ``$\mathbf{+2.98\%}$'' ``$\mathbf{+13.76\%}$'', and ``$\mathbf{+1.33\%}$'' improvement in Rank-1, Rank-5, Rank-10, mAP, and mINP metrics, respectively.  Adding CIDHL and MBSOS individually also shows performance improvements in most metrics. In Mode 2, the AGW algorithm with the addition of both CIDHL and MBSOS also achieves the best performance, showing ``$\mathbf{+17.72\%}$'', ``$\mathbf{+4.36\%}$'', ``$\mathbf{+1.95\%}$'', ``$\mathbf{+16.24\%}$'', and ``$\mathbf{+2.67\%}$'' improvements in Rank-1, Rank-5, Rank-10, mAP, and mINP metrics, respectively. Adding CIDHL and MBSOS individually also shows performance improvements in most metrics, too. The performance of other algorithms is similar to AGW.

\begin{table}[]
\caption{Comparison to the state-of-the-art methods on SYSU-MM01 datasets in mode 2.}
\label{tab:4}
\begin{tabular}{c|ccccc}
\hline
\multirow{2}{*}{Method}       & \multicolumn{5}{c}{Mode 2}    \\ \cline{2-6} 
                              & Rank-1 & Rank-5 & Rank-10 & mAP &mINP \\ \hline
DART $w/o$ mix       &60.99   &89.07    &95.59    &68.34 &64.07   \\
DART                 &84.62   &96.84    &98.73    &51.78 &12.09   \\ 
LUPI $w/o$ mix       &49.95   &86.12    &91.46    &55.49 &48.18   \\
LUPI                 &49.14   &76.57    &85.58    &30.05 &5.63   \\ \hline
AGW $w/o$ mix         &55.66   &84.42    &92.48&42.32 &5.82  \\
AGW         &73.09   &93.91    &97.52&37.00 &2.83  \\
AGW $+$ M                     &73.43\textcolor{red}{$\uparrow$}   &94.18\textcolor{red}{$\uparrow$}    &97.61\textcolor{red}{$\uparrow$}&37.15\textcolor{red}{$\uparrow$} &2.83  \\
AGW $+$ C                     &90.36\textcolor{red}{$\uparrow$}   &98.12\textcolor{red}{$\uparrow$}    &99.31\textcolor{red}{$\uparrow$}&53.21\textcolor{red}{$\uparrow$}    &5.32\textcolor{red}{$\uparrow$}    \\
AGW $+$ C $+$ M               &\textbf{90.81}\textcolor{red}{$\uparrow$}   &\textbf{98.27}\textcolor{red}{$\uparrow$}    &\textbf{99.37}\textcolor{red}{$\uparrow$}&\textbf{53.24}\textcolor{red}{$\uparrow$}    &\textbf{5.50}\textcolor{red}{$\uparrow$}    \\ \hline
HCTL $w/o$ mix &73.46   &90.30    &95.74&59.74   &17.95     \\
HCTL &67.36   &90.01    &94.69&33.94   &2.79     \\
HCTL $+$ M                    &68.91\textcolor{red}{$\uparrow$}   &90.96\textcolor{red}{$\uparrow$}    &95.14\textcolor{red}{$\uparrow$}&34.89\textcolor{red}{$\uparrow$}   &2.98\textcolor{red}{$\uparrow$}      \\
HCTL $+$ C                    &87.74\textcolor{red}{$\uparrow$}   &97.61\textcolor{red}{$\uparrow$}    &99.16\textcolor{red}{$\uparrow$}&47.16\textcolor{red}{$\uparrow$}    &4.33\textcolor{red}{$\uparrow$}     \\
HCTL $+$ C $+$ M              &\textbf{88.28}\textcolor{red}{$\uparrow$}   &\textbf{97.67}\textcolor{red}{$\uparrow$}    &\textbf{99.25}\textcolor{red}{$\uparrow$}&\textbf{47.45}\textcolor{red}{$\uparrow$}     &\textbf{4.39}\textcolor{red}{$\uparrow$}    \\\hline
DEEN $w/o$ mix &72.01   &94.79  &98.01  &55.51&14.59         \\
DEEN &67.09   &87.44    &93.32&11.42    &1.36     \\
DEEN $+$ M                    &61.01   &83.66    &90.72&30.93\textcolor{red}{$\uparrow$}    &4.53\textcolor{red}{$\uparrow$}      \\
DEEN $+$ C                    &81.89\textcolor{red}{$\uparrow$}   &\textbf{95.58}\textcolor{red}{$\uparrow$}    &\textbf{98.00}\textcolor{red}{$\uparrow$} &49.77\textcolor{red}{$\uparrow$}    &5.14\textcolor{red}{$\uparrow$}     \\
DEEN $+$ C $+$ M              &\textbf{82.52}\textcolor{red}{$\uparrow$}   &95.50\textcolor{red}{$\uparrow$}    &97.85\textcolor{red}{$\uparrow$}  &\textbf{50.04}\textcolor{red}{$\uparrow$}   &\textbf{5.44}\textcolor{red}{$\uparrow$}      \\ \hline
\end{tabular}
\end{table}

\begin{table}[]
\caption{Comparison to the state-of-the-art methods on LLCM datasets in mode 1.}
\label{tab:5}
\begin{tabular}{c|ccccc}
\hline
\multirow{2}{*}{Method}       & \multicolumn{5}{c}{Mode 1}    \\ \cline{2-6} 
                              & Rank-1 & Rank-5 & Rank-10 & mAP &mINP \\ \hline
AGW $w/o$ mix        &57.49   &79.46    &86.37&44.40 &16.51  \\
AGW         &51.15   &68.09    &74.53&36.04 &10.49  \\
AGW $+$ M                     &51.15   &68.10\textcolor{red}{$\uparrow$}    &74.53&36.04 &10.49  \\
AGW $+$ C                     &\textbf{59.51}\textcolor{red}{$\uparrow$}   &\textbf{75.07}\textcolor{red}{$\uparrow$}    &\textbf{80.82}\textcolor{red}{$\uparrow$}&\textbf{42.02}\textcolor{red}{$\uparrow$}   &\textbf{10.80}\textbf{}     \\
AGW $+$ C $+$ M               &\textbf{59.51}\textcolor{red}{$\uparrow$}   &\textbf{75.07}\textcolor{red}{$\uparrow$}    &\textbf{80.82}\textcolor{red}{$\uparrow$}&\textbf{42.02}\textcolor{red}{$\uparrow$} &\textbf{10.80}\textcolor{red}{$\uparrow$} \\ \hline
HCTL $w/o$ mix &44.84  &68.21    &77.32&29.16  &4.09      \\
HCTL &37.68   &55.54    &62.61&22.74    &4.20    \\
HCTL $+$ M                    &\textbf{37.70}\textcolor{red}{$\uparrow$}   &\textbf{55.56}\textcolor{red}{$\uparrow$}    &62.59&\textbf{22.74}\textcolor{red}{$\uparrow$}     &4.20    \\
HCTL $+$ C                    &35.36   &55.14    &\textbf{63.08}\textcolor{red}{$\uparrow$}&22.54    &\textbf{4.30}\textcolor{red}{$\uparrow$}     \\
HCTL $+$ C $+$ M              &35.36   &55.14    &\textbf{63.08}\textcolor{red}{$\uparrow$}&22.54  &\textbf{4.30}\textcolor{red}{$\uparrow$} \\\hline
DEEN $w/o$ mix &70.69   &86.60    &91.05&55.62   &21.96      \\
DEEN  &58.99   &74.39    &79.91&43.43   &13.78      \\
DEEN $+$ M                    &58.18   &73.69    &79.55&43.12    &\textbf{14.06}\textcolor{red}{$\uparrow$}      \\
DEEN $+$ C                    &\textbf{61.49}\textcolor{red}{$\uparrow$}   &\textbf{76.22}\textcolor{red}{$\uparrow$}    &\textbf{81.23}\textcolor{red}{$\uparrow$} &\textbf{45.35}\textcolor{red}{$\uparrow$}  &13.19       \\
DEEN $+$ C $+$ M              &61.08\textcolor{red}{$\uparrow$}   &75.85\textcolor{red}{$\uparrow$}    &80.98\textcolor{red}{$\uparrow$}  &45.22\textcolor{red}{$\uparrow$}   &13.49     \\ \hline
\end{tabular}
\end{table}

\begin{table}[]
\caption{Comparison to the state-of-the-art methods on LLCM datasets in mode 2.}
\label{tab:6}
\begin{tabular}{c|ccccc}
\hline
\multirow{2}{*}{Method}       & \multicolumn{5}{c}{Mode 2}    \\ \cline{2-6} 
                              & Rank-1 & Rank-5 & Rank-10 & mAP &mINP \\ \hline
AGW $w/o$ mix        &57.51   &79.46    &86.39&44.40 &16.51  \\
AGW         &53.60   &71.60    &77.80&37.33  &9.39 \\
AGW $+$ M                     &53.60\textcolor{red}{$\uparrow$}   &71.60\textcolor{red}{$\uparrow$}    &77.80\textcolor{red}{$\uparrow$}&37.33\textcolor{red}{$\uparrow$} &9.39  \\
AGW $+$ C                     &\textbf{61.51}\textcolor{red}{$\uparrow$}   &\textbf{76.72}\textcolor{red}{$\uparrow$}    &\textbf{81.86}\textcolor{red}{$\uparrow$}&\textbf{42.13}\textcolor{red}{$\uparrow$}     &\textbf{9.78}\textcolor{red}{$\uparrow$}   \\
AGW $+$ C $+$ M               &\textbf{61.51}\textcolor{red}{$\uparrow$}   &\textbf{76.72}\textcolor{red}{$\uparrow$}    &\textbf{81.86}\textcolor{red}{$\uparrow$}&\textbf{42.13}\textcolor{red}{$\uparrow$} &\textbf{9.78}\textcolor{red}{$\uparrow$}  \\ \hline
HCTL $w/o$ mix &48.42   &74.04    &83.14&33.26    &8.71    \\
HCTL  &38.97   &58.12    &66.32&23.58  &4.92      \\
HCTL $+$ M                    &38.31   &56.93    &64.20&22.81   &3.71      \\
HCTL $+$ C                    &38.45   &\textbf{59.15}\textcolor{red}{$\uparrow$}    &68.04\textcolor{red}{$\uparrow$}&\textbf{24.06}\textcolor{red}{$\uparrow$}  &\textbf{5.42}\textcolor{red}{$\uparrow$}       \\
HCTL $+$ C $+$ M              &38.94   &59.12\textcolor{red}{$\uparrow$}    &\textbf{68.94}\textcolor{red}{$\uparrow$}&22.96\textcolor{red}{$\uparrow$}    &4.00     \\\hline
DEEN $w/o$ mix &70.71   &86.60    &91.05&55.62   &21.95      \\
DEEN  &62.91   &77.53    &82.69&46.10    &12.74     \\
DEEN $+$ M                    &62.93\textcolor{red}{$\uparrow$}   &77.09    &82.35&46.15\textcolor{red}{$\uparrow$}     &\textbf{13.04}     \\
DEEN $+$ C                    &\textbf{65.06}\textcolor{red}{$\uparrow$}   &\textbf{78.28}\textcolor{red}{$\uparrow$}    &\textbf{83.73}\textcolor{red}{$\uparrow$} &46.87\textcolor{red}{$\uparrow$}     &\textbf{13.13}    \\
DEEN $+$ C $+$ M              &64.70\textcolor{red}{$\uparrow$}   &78.15\textcolor{red}{$\uparrow$}    &83.49\textcolor{red}{$\uparrow$}  &\textbf{46.90}\textcolor{red}{$\uparrow$}     &\textbf{13.38}\textcolor{red}{$\uparrow$}    \\ \hline
\end{tabular}
\end{table}

\textbf{Comparison on LLCM.}
As can be seen in Tables~\ref{tab:3} and~\ref{tab:4}, a general improvement in performance occurs with the addition of our method on SYSU-MM01, too. In Mode 1, the AGW algorithm achieves the best performance with the addition of both CIDHL and MBSOS, showing ``$\mathbf{+8.36\%}$'', ``$\mathbf{+6.97\%}$'', ``$\mathbf{+6.29\%}$'' ``$\mathbf{+5.98\%}$'', and ``$\mathbf{+0.31\%}$'' improvement in Rank-1, Rank-5, Rank-10, mAP, and mINP metrics, respectively.  Adding CIDHL and MBSOS individually also shows performance improvements in most metrics. In Mode 2, the AGW algorithm with the addition of both CIDHL and MBSOS also achieves the best performance, showing ``$\mathbf{+7.91\%}$'', ``$\mathbf{+5.12\%}$'', ``$\mathbf{+4.06\%}$'' ``$\mathbf{+4.80\%}$'', and ``$\mathbf{+0.39\%}$'' improvements in Rank-1, Rank-5, Rank-10, mAP, and mINP metrics, respectively. Adding CIDHL and MBSOS individually also shows performance improvements in most metrics, too. The performance of other algorithms is similar to AGW.

\begin{table}[!htbp]
\caption{Ablation study of $\delta$. The \textbf{bold} is the best result on SYSU-MM01 dataset in mode 1.}
\label{tab:7_1}
\centering
\begin{tabular}{cl|ccccc}
\hline
\multicolumn{2}{c|}{\multirow{2}{*}{$\delta$}} & \multicolumn{5}{c}{Mode 1}                                                   \\ \cline{3-7} 
\multicolumn{2}{c|}{}                      & Rank-1         & Rank-5         & Rank-10        & mAP            & mINP          \\ \hline
\multicolumn{2}{c|}{Origin}                & 70.22          & 91.76          & 95.99          & 34.78          & 1.97          \\ \hline
\multicolumn{2}{c|}{0}                     & 55.73          & 78.25          & 86.76          & 6.42           & 1.15          \\ \hline
\multicolumn{2}{c|}{0.1}                   & 85.68          & 96.58          & 98.33          & 40.13          & 1.36          \\ \hline
\multicolumn{2}{c|}{0.2}                   & \textbf{88.76} & \textbf{97.36} & \textbf{98.84} & \textbf{48.79} & \textbf{3.29} \\ \hline
\multicolumn{2}{c|}{0.3}                   & 77.85          & 93.37          & 96.30          & 38.88          & 2.73          \\ \hline
\multicolumn{2}{c|}{0.4}                   & 85.58          & 96.75          & 98.73          & 41.55          & 1.58          \\ \hline
\multicolumn{2}{c|}{0.5}                   & 83.30          & 94.93          & 97.68          & 42.70          & 2.68          \\ \hline
\end{tabular}
\end{table}

\begin{table}[!htbp]
\caption{Ablation study of $\delta$. The \textbf{bold} is the best result on SYSU-MM01 dataset in mode 2.}
\label{tab:7_2}
\centering
\begin{tabular}{cl|ccccc}
\hline
\multicolumn{2}{c|}{\multirow{2}{*}{$\delta$}} & \multicolumn{5}{c}{Mode 2}                                                    \\ \cline{3-7} 
\multicolumn{2}{c|}{}                      & Rank-1         & Rank-5         & Rank-10        & mAP            & mINP          \\ \hline
\multicolumn{2}{c|}{Origin}                & 73.09          & 93.91          & 97.52          & 37.00          & 2.83          \\ \hline
\multicolumn{2}{c|}{0}                     & 59.76          & 82.13          & 89.74          & 9.97           & 1.32          \\ \hline
\multicolumn{2}{c|}{0.1}                   & 88.04          & 97.49          & 99.11          & 44.73          & 2.42          \\ \hline
\multicolumn{2}{c|}{0.2}                   & \textbf{90.36} & \textbf{98.12} & \textbf{99.31} & \textbf{53.21} & \textbf{5.32} \\ \hline
\multicolumn{2}{c|}{0.3}                   & 84.58          & 95.61          & 97.88          & 43.24          & 4.05          \\ \hline
\multicolumn{2}{c|}{0.4}                   & 85.56          & 97.88          & 98.81          & 44.82          & 2.96          \\ \hline
\multicolumn{2}{c|}{0.5}                   & 86.28          & 97.37          & 98.93          & 48.07          & 4.67          \\ \hline
\end{tabular}
\end{table}

\subsection{Ablation Study}
\label{Ablation}
We adopted the AGW method to perform our ablation experiments on the SYSU-MM01 dataset in Rank-1, Rank-5, Rank-10, and mAP metrics. For length limitation, we only show the ablation experiments for the $\delta$ parameter in Eq.5, the ablation experiments for the $\lambda$ parameter in Eq.7 and Eq.8 are shown in the supplementary material, and the effects of the addition of the method are shown in the previous tables.

\begin{table}[]
\caption{Ablation study of $\lambda$. The \textbf{bold} is the best result on SYSU-MM01 dataset in mode 1.}
\label{tab:8_1}
\centering
\begin{tabular}{cl|ccccc}
\hline
\multicolumn{2}{c|}{\multirow{2}{*}{$\lambda$}} & \multicolumn{5}{c}{Mode 1}                                                  \\ \cline{3-7} 
\multicolumn{2}{c|}{}                      & Rank-1         & Rank-5         & Rank-10        & mAP            & mINP          \\ \hline
\multicolumn{2}{c|}{0.99}                  & \textbf{70.73} & \textbf{92.02} & 96.09          & 34.76          & 1.96          \\ \hline
\multicolumn{2}{c|}{0.999}                 & 70.26          & 91.76          & 96.01          & \textbf{34.78} & 1.97          \\ \hline
\multicolumn{2}{c|}{0.9999}                & 70.22          & 91.74          & 95.99          & \textbf{34.78} & 1.97          \\ \hline
\multicolumn{2}{c|}{1}                     & 70.22          & 91.76          & 95.99          & \textbf{34.78} & 1.97          \\ \hline
\multicolumn{2}{c|}{1.001}                 & 69.90          & 91.49          & \textbf{98.86} & \textbf{34.78} & 1.96          \\ \hline
\multicolumn{2}{c|}{1.01}                  & 70.12          & 91.72          & 95.95          & 34.75          & \textbf{1.98} \\ \hline
\multicolumn{2}{c|}{1.1}                   & 62.89          & 86.31          & 92.54          & 31.58          & 1.94          \\ \hline
\end{tabular}
\end{table}

\begin{table}[]
\caption{Ablation study of $\lambda$. The \textbf{bold} is the best result on SYSU-MM01 dataset in mode 2.}
\label{tab:8_2}
\centering
\begin{tabular}{cl|ccccc}
\hline
\multicolumn{2}{c|}{\multirow{2}{*}{$\lambda$}} & \multicolumn{5}{c}{Mode 2}                                                   \\ \cline{3-7} 
\multicolumn{2}{c|}{}                      & Rank-1         & Rank-5         & Rank-10        & mAP            & mINP          \\ \hline
\multicolumn{2}{c|}{0.99}                  & \textbf{73.43} & \textbf{94.18} & \textbf{97.61} & \textbf{37.15} & \textbf{2.83} \\ \hline
\multicolumn{2}{c|}{0.999}                 & 73.09          & 93.94          & 97.55          & 37.02          & \textbf{2.83} \\ \hline
\multicolumn{2}{c|}{0.9999}                & 73.09          & 93.94          & 97.52          & 37.00          & \textbf{2.83} \\ \hline
\multicolumn{2}{c|}{1}                     & 73.09          & 93.91          & 97.50          & 37.00          & \textbf{2.83} \\ \hline
\multicolumn{2}{c|}{1.001}                 & 72.61          & 93.47          & 97.34          & 36.74          & \textbf{2.83} \\ \hline
\multicolumn{2}{c|}{1.01}                  & 73.03          & 93.88          & 97.52          & 36.97          & 2.82          \\ \hline
\multicolumn{2}{c|}{1.1}                   & 64.46          & 85.95          & 92.81          & 31.25          & 2.61          \\ \hline
\end{tabular}
\end{table}

As can be seen in Table~\ref{tab:7_1} and~\ref{tab:7_2}, in Mode 1, compared with the performance without our CIDHL, there are at least ``$\mathbf{+7.63\%}$'', ``$\mathbf{+1.61\%}$'', ``$\mathbf{+0.31\%}$'', ``$\mathbf{+4.10\%}$'', and ``$\mathbf{+0.76\%}$'' performance improvements in Rank-1, Rank-5, Rank-10, mAP, and mINP metrics ($\delta$ is set to 0.3), and at most ``$\mathbf{+18.54\%}$'', ``$\mathbf{+5.60\%}$'', ``$\mathbf{+2.85\%}$'', ``$\mathbf{+14.01\%}$'', and ``$\mathbf{+1.32\%}$'' performance improvements ($\delta$ is set to $0.2$), respectively, when our CIDHL is added. ($\delta$ is set to 0.2). In Model 2, compared to the performance without our CIDHL, there is at least ``$\mathbf{+11.49\%}$'', ``$\mathbf{+1.70\%}$'', ``$\mathbf{+0.36\%}$'', ``$\mathbf{+6.24\%}$'', and ``$\mathbf{+1.22\%}$'' performance improvement ($\delta$ is set to $0.3$) and at most ``$\mathbf{+17.27\%}$'', ``$\mathbf{+4.21\%}$'', ``$\mathbf{+1.79\%}$'', ``$\mathbf{+16.21\%}$'', and ``$\mathbf{+2.49\%}$'' performance improvement ($\delta$ is set to 0.2) in the four metrics of Rank-1, Rank-5, Rank-10, and mAP, respectively, with the addition of our CIDHL. Besides, in Mode 1, compared to only $L_{cid}$ used ($\delta$ is set to $0$), the whole $L_{CIDHL}$ achieves the best performance, showing ``$\mathbf{+33.03\%}$'', ``$\mathbf{+19.11\%}$'', ``$\mathbf{+12.08\%}$'' ``$\mathbf{+42.37\%}$'', and ``$\mathbf{+2.14\%}$'' improvement in Rank-1, Rank-5, Rank-10, mAP, and mINP metrics, respectively. In Mode 2, compared to only $L_{cid}$ used ($\delta$ is set to 0), the whole $L_{CIDHL}$ achieves the best performance, showing ``$\mathbf{+30.60\%}$'', ``$\mathbf{+15.99\%}$'', ``$\mathbf{+9.57\%}$'' ``$\mathbf{+43.24\%}$'', and ``$\mathbf{+4.00\%}$'' improvement in Rank-1, Rank-5, Rank-10, mAP, and mINP metrics, respectively.



As can be seen in Table~\ref{tab:8_1} and~\ref{tab:8_2}, both in Mode 1 and Mode 2, the addition of our MBSOS brings different effects with different distance adjustment ratios, and for most of the metrics, the best results are achieved when $\lambda$ is set to $0.99$.

\begin{figure*}[!t]
\centering
\includegraphics[width=0.95\textwidth, height=0.6\textwidth]{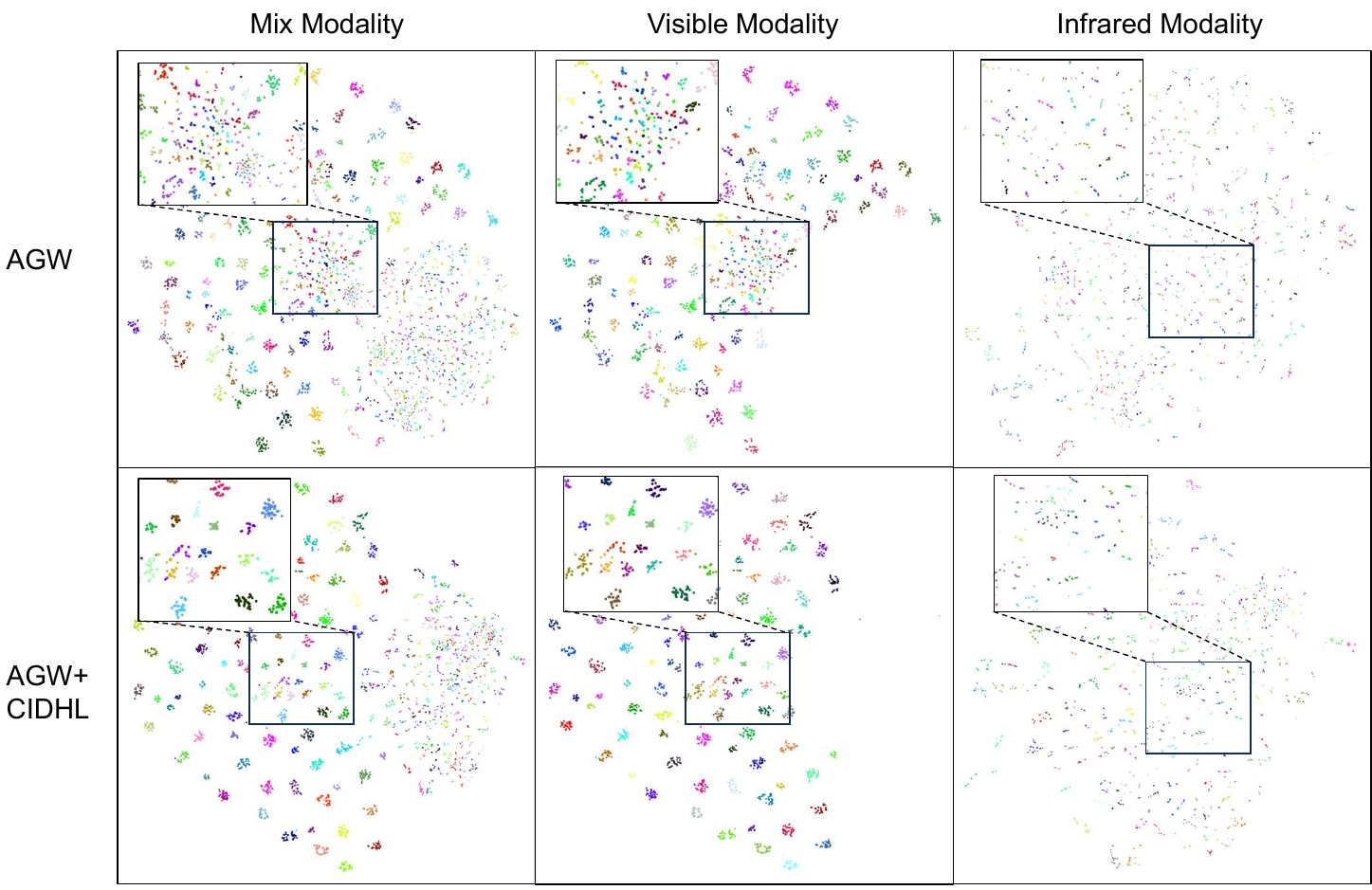}
\caption{The t-SNE Visualization of only AGW (first row) and AGW addition with our CIDHL (second row). (a) Results of mix modality. (b) Results of Visible modality. (b) Results of infrared modality. Different colors represent different identities, dots for visible modal samples, and stars for infrared modal samples.}
\label{fig:five}
\end{figure*}

\subsection{Visualization}
\label{Visualization}
As illustrated in Figure~\ref{fig:five}, we compared the t-SNE visualization results of the original AGW and the AGW with the addition of our CIDHL loss. After adding our CIDHL, it is clear from Figure~\ref{fig:five} (a) that the samples of the two modalities are well differentiated, and from Figure~\ref{fig:five} (b), (c) that the same-identity samples within each modality are more aggregated, and the different-identity samples are easier to distinguish. The above phenomenon combined with the improved performance in the comparison experiments is a good proof of its effectiveness in solving the modality confusion problem.

\section{Conclusion}
In response to the limited application scenarios of the visible infrared bi-modality mutual retrieval paradigm in existing \textbf{V}isible-\textbf{I}nfrared person re-identification (VI-ReID), this paper proposes a new and practical mix-modality retrieval paradigm. In the mix-modality retrieval paradigm, the query set and the gallery set are both collections of the mixture of visible and infrared modalities. Based on the new mix-modality paradigm, we propose a \textbf{M}ix-\textbf{M}odality person re-identification (MM-ReID) task and explore the effect of different modality mixing ratios on the performance of the model, and construct the corresponding mix-modality test sets for the existing datasets. In MM-ReID there is a problem of confusing cross-modality identity matching due to identity-independent information such as similar colors in the same modality, we summarize this problem as the modality confusion problem. To address this modality confusion problem, we propose a \textbf{C}ross-\textbf{I}dentity \textbf{D}iscrimination \textbf{H}armonization \textbf{L}oss (CIDHL) and a \textbf{M}odality \textbf{B}ridge \textbf{S}imilarity \textbf{O}ptimization \textbf{S}trategy (MBSOS). The former pulls the centers of samples with the same identity closer together and pushes the centers of samples with different identities farther apart, while at the same time aggregating the distances from samples with the same identity in the same modality to the sample centers. The latter optimizes the similarity distance metric from the query point to the queried point by finding bridge samples in the gallery. Extensive experiments demonstrate the transferability and ability to cope with modality confusion of our methods. The code and datasets will be released after the paper is accepted.

\section{Limitation}
Although this article has explored the new proposed mix-modality person re-identification task to some extent, there are still many areas where more effort needs to be invested in refining it. Among them, the most important one is the new way of data utilization during training. Unlike the existing two-branch model which inputs two modality images separately, the design of the new mix-modality inputs and the corresponding model will allow the model to learn ways to cope with modality confusion more efficiently during training. In addition, the improvement of the proposed modality bridge strategy is average, in addition to utilizing the shortest paths, exploring the potential relationship between the same-modality sample distances and cross-modality sample distances may be more helpful for the final performance. The above is the focus of our future work.

\begin{acks}
This work was supported by the National Nature Science Foundation of China No.62376201.
\end{acks}

\bibliographystyle{ACM-Reference-Format}
\bibliography{ref}
\end{document}